\documentclass{WileyMSP-template}
\usepackage{xcolor}
\usepackage{algorithm}
\usepackage{algpseudocode}
\usepackage{amssymb}
\usepackage{microtype}
\usepackage{bm, amsmath}
\usepackage[font=sf]{caption}
\usepackage[normalem]{ulem}

\usepackage[skip=0.75\baselineskip plus 2pt]{parskip}
\newcommand{\link}{\mathcal{L}k}

\begin{document}

\pagestyle{fancy}
\rhead{\includegraphics[width=2.5cm]{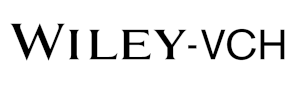}}

\title{Stochastic Entanglement of Deterministic Origami Tentacles For Robust Robotic Grasping}
\maketitle

\author{Alec Boron$^+$}
\author{Bokun Zheng$^+$}
\author{Ziyang Zhou$^+$}
\author{Noel Naughton}
\author{Suyi Li*}

\begin{affiliations}

Department of Mechanical Engineering, Virginia Tech, VA, USA\\[6pt]
*corresponding author, email address: \texttt{suyili@vt.edu}\\[6pt]
$^+$A.B., B.Z., and Z.Z. contributed equally to this work.

\end{affiliations}


\begin{abstract}

Origami-inspired robotic grippers have shown promising potential for object manipulation tasks due to their compact volume and mechanical flexibility. However, robust capture of objects with random shapes in dynamic working environments often comes at the cost of additional actuation channels and control complexity. Here, we introduce a tendon-driven, robust origami tentacle gripper by exploiting a synergy between local, deterministic deformation programming and global, stochastic entanglements. Each tentacle features carefully placed holes (for routing an actuation tendon), origami creases, and a tapered shape.  By tailoring these design features, one can prescribe the shrinking, bending, and twisting deformation, eventually creating deterministic coiling with a simple tendon pull.  Then, when multiple coiling tentacles are placed in proximity, stochastic entanglement emerges, allowing the tentacles to braid, knot, and grip objects with random shapes. We derived a simulation model by integrating origami mechanics with Cosserat rods to correlate origami design, tentacle deformation, and collective grasping performance. Then, we experimentally tested how these entangling origami tentacles can grasp objects under gravity and in water. A stow-and-release deployment mechanism was also tested to simulate in-orbit grasping. Overall, this entanglement-enabled tentacle gripper presents a route toward robust object grasping with simple design and actuation.

\end{abstract}

\section{Introduction}

Over the past decades, origami folding has evolved from a recreational art to an engineering framework for developing thin sheet materials into deployable mechanisms with versatile 3D shapes and functionalities~\cite{wang2026embodying, rus2018design}. A key driving force underpinning this evolution is the unique ability to prescribe (or ``program'') the origami's shape transformation by tailoring the crease geometry. That is, folding deformation typically concentrates along prescribed crease lines, while the facets in between the creases remain mostly undeformed (even rigid in some cases)~\cite{peng2025thick, chen2015origami}. As a result, origami is \emph{kinematically deterministic and programmable}, meaning that the geometric layout of the crease pattern dictates the transformation in its 3D shapes and the evolution of its mechanical properties during folding~\cite{bershadsky2026digital}.
Foundational rigid-folding or structural-mechanics models can capture these behaviors by idealizing the folded origami as a linkage mechanism~\cite{greenberg2011identifying} or truss-frame network~\cite{li2019architected}. They can analyze kinematics and mechanics directly from crease geometry and constitutive material properties~\cite{filipov2017bar, woodruff2020bar, zhu2022review}. 
Customizing the crease patterns offers us a large design space. This, combined with the model's evolution, has allowed us to engineer a wide array of origami-based deployable structures~\cite{filipov2015origami, zhou2025hyper, deshpande2024golden}, functional materials~\cite{ning2018assembly,wei2025biodegradable,wang2026nanoasperity}, soft electronics~\cite{qiu2026seal}, and mechanical computers~\cite{zhou2024self,chen2025general}. 

\medskip

The kinematically deterministic and programmable nature of origami has also fostered the creation of soft robotic grippers~\cite{zhang2025application}. The idea is to use folding to convert compact sheets or shells into a three-dimensional enclosure for object grasping and manipulation. By customizing the underlying crease design and integrating novel actuation methods such as fluidics~\cite{li2019vacuum, chen2021soft, cao2024design}, pulled tendons~\cite{liu2024tendon}, responsive materials~\cite{miyashita2017robotic, tolley2014self}, or hybrid schemes~\cite{cao2024design}, one can uniquely design soft grippers for specific target objects and task environments~\cite{ye2026integrating, geckeler2022bistable}. In addition, elastic instability and snap-through can further amplify the speed and magnitude of origami folding for rapid, non-invasive grasping \cite{hong2022boundary, yang2021grasping}.

\medskip
However, the kinematically deterministic nature of origami can also significantly constrain grasping functionality. Once the crease pattern is defined, the corresponding folding trajectories and grasp modes are fixed. As a result, robotic grasping is typically demonstrated on a small set of curated targets, rather than a truly open-ended task environment \cite{shintake2018soft, firouzeh2017under}. Extending the grasping capability to diverse object sizes and shapes often requires more complex origami designs and additional actuation hardware, such as multi-chamber pneumatics, valve networks, variable-stiffness elements, or snapping mechanisms, increasing the overall complexity \cite{hong2025reprogrammable}. Even though fabrication methods have improved significantly via monolithic manufacturing, practical constraints still remain~\cite{zhai2023desktop, wu2025monolithic}. These limitations motivated us to adopt \emph{a new grasping strategy that is robust enough to grasp objects with radically different shapes under diverse working conditions, while maintaining the simplicity of origami design and actuation.}

\medskip
Therefore, in this study, we introduce \emph{stochastic entanglement behaviors} into the origami mechanism as a novel approach toward robust and versatile object manipulation. Arrays of compliant filaments and tentacle-like appendages have been shown to achieve remarkably reliable object capture through contact-driven coiling and braiding, where grasp emerges from redundant contacts and topological locking rather than deterministic enclosure alone~\cite{becker2022active, li2024pegrip}. Prior works in spiral and twisting grippers further demonstrated that curvature and torsion in the body can enlarge the capture envelope and improve retention across different targets \cite{wang2025spirobs, yang2020twining}. Together, these studies suggest that entanglement can mitigate the limitations of conventional origami grippers by shifting the objective from accurately executing a planned folding motion to robustly harnessing collective behaviors.

\begin{figure}[b!]
    \centering
    \vspace{-10pt}
    \includegraphics[scale=0.9]{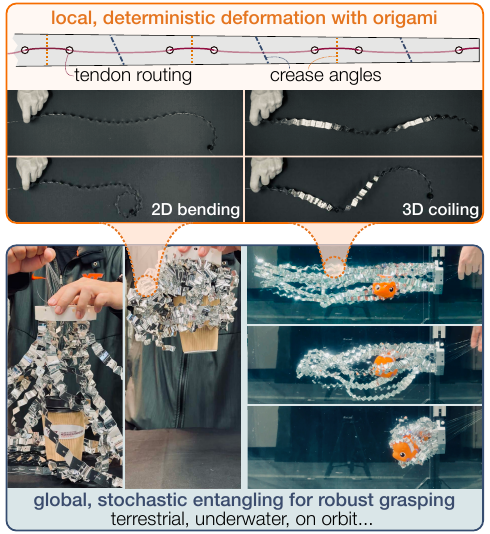}
    \vspace{-5pt}
    \caption{\textbf{Overview}. A synergy between local, deterministic origami tentacle deformation and global, stochastic entangling enables robust robotic grasping in air, underwater, and even in orbit.}
    \label{fig:overview}
\end{figure}

\medskip
Building on this insight, we design and validate a lightweight, easy-to-fabricate robotic platform consisting of multiple tendon-driven origami ``tentacles'' that can achieve collective entanglement for robust object grasping.
Each origami tentacle---named for its deformation behavior resembling cephalopod appendages~\cite{tekinalp2024topology}---is a thin, tapered sheet with an array of accordion folds. These folds can be programmed to \emph{deterministically} create 3D coiling deformation under a global tendon pulling. When multiple tentacles operate in parallel and in proximity, they entangle \emph{stochastically}, wrapping and knotting around the target object to secure a robust grip. 
To the best of our knowledge, \textit{such an entanglement-enabled grasping approach is the first attempt in robotics to achieve synergy between local, deterministic deformation programming (at the single-tentacle level) and global, stochastic emergent behaviors (at the multiple-tentacle level)} (Figure \ref{fig:overview}).

\medskip
In what follows, we detail the design and deformation programming of a single origami tentacle, and then analyze the collective grasping behavior in air (with gravity), underwater (with buoyancy and drag), and a simulated space environment (without gravity). Overall, this study presents a comprehensive framework --- origami design, mechanics modeling, dynamic simulation, and fabrication --- that opens a practical pathway toward compact and robust object grasping under uncertain and complex task conditions.

\section{Deterministic Deformation Programming of Single Tentacles}

\begin{figure}[t!]
    \centering
    \includegraphics[width=\textwidth]{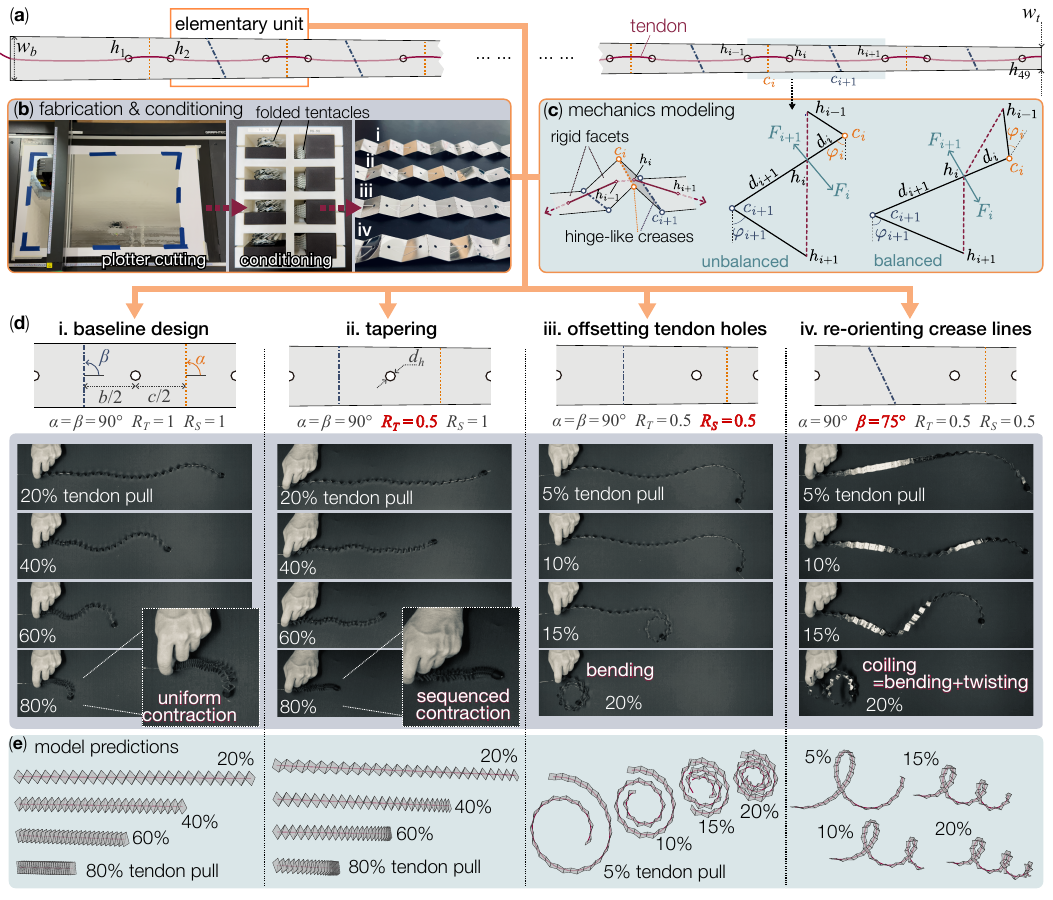}
    \vspace{-20pt}
    \caption{\textbf{Design, modeling, and experimental characterization of a single origami tentacle.} 
    (a) The design of an origami tentacle. The actuation tendon weaves through the holes $h_1$ to $h_{49}$ and is fixed to the tentacle at the tip. An elementary unit, which contains three holes and two creases, is highlighted.
    (b) Fabrication includes accurate tentacle cutting from a plotter and a pre-conditioning step, where fully folded tentacles are held in a container for eight hours to release residual stress (similar to annealing). 
    (c) Mechanics modeling. A force balance is performed on all elementary units to identify the folding angles $\{\phi_i | i=1..N\}$ that produce force equilibrium at all tendon-contact holes $h_i$. On the left, a unit is shown with a hypothetical configuration where the bending torques developed by the two creases are equal ($\tau_i=\tau_{i+1}$) and all holes are aligned so the tendon runs through them in a straight line. In this case, the moment arms differ ($d_i \neq d_{i+1}$), yielding a mismatch between the normal forces, which tendon-hole point $h_i$ cannot sustain due to the tendon's compliance, making this configuration mechanically unstable.
    On the right, a configuration where force equilibrium is achieved is shown. Here, the crease angles $\phi_i$ are adjusted such that the normal forces that develop at hole $h_i$ are balanced. This yields an equilibrium configuration with a bend in the tendon routing, which in turn leads to the emergence of tentacle coiling when multiple elementary units are concatenated together. 
    (d) The deterministic deformation programming. The top figure in each column shows the design of an elementary unit; meanwhile, the pictures below are high-speed snapshots of the corresponding origami tentacle deformation (Movie S1). Here, the ``\% tendon pull'' is the ratio of retracted tendon length over the initial tentacle length. 
    (e) The corresponding mechanics model simulation shows reasonable agreement.}
    \label{fig:single}
\end{figure}

A single origami tentacle is fabricated as a long, trapezoidal ribbon cut out of a 0.1 mm-thick Mylar (PET) sheet on a plotter cutter (Graphtec Cutting Pro FCX4000-60). Mylar is compliant, durable, and tear-resistant, allowing for repeatable deformation outcomes even after many folding/unfolding cycles. Each tentacle has an overall length $l$, base width $w_b$, and tip width $w_t$, so we define a ``tapering ratio'' $R_T=w_t/w_b$ to describe its overall shape (Figure~\ref{fig:single}a). Along the centerline of the tentacle, we use the plotter to cut circular holes $h_{1}$--$h_{49}$, each with a diameter $d_h$. An actuation tendon can ``weave'' through these holes on both sides of the tentacle until it is fixed to the tentacle at the tip. Then, we use the plotter cutter to perforate origami crease lines in between every pair of tendon holes. This way, an ``elementary unit'' of the origami tentacle consists of three tendon holes and two crease lines, and the whole tentacle's design is a repeated array of these units. Within a unit, the distances between the tendon holes are $b$ and $c$, so we can define a ``spacing ratio'' $R_S=b/c$. The crease lines are placed at equal distances between adjacent holes, and their orientations are defined by $\alpha$ and $\beta$ angles, respectively (Figure~\ref{fig:single}d).  
Our central idea is that by simply pulling the tendon, one can shorten the tentacle and fold the origami creases within, thus generating a distributed deformation. And the deformation characteristics can be uniquely determined by the tapering ratio $R_T$, spacing ratio $R_S$, and crease angles $\alpha$ and $\beta$.

\subsection{Correlating Tentacle Deformation to Origami Design}

To correlate the tentacle's deformation characteristics with these key geometric parameters. We fabricated a set of four tentacle prototypes (Figure~\ref{fig:single}b,d). They all have the same overall length $l=750$~mm, base width $w=20$~mm, tendon hole diameter $d_h=2$~mm, and unit length $b+c=30$~mm, but other design parameters vary. These tentacles are precisely cut out of a Mylar sheet on a plotter, manually folded along the crease lines to maximum deformation, and then held in a 3D-printed mold for 8 hours before testing.  This final ``conditioning step'' is crucial because it helps release residual crease stresses introduced during fabrication, making the tentacle deformation consistent across different samples (aka. minimal fabrication variation). For this reason, \textit{all} tentacles used throughout this study are conditioned before testing (more details in Methods and SI Section 1).

\textbf{A uniform and symmetric design generates linear contraction}:
The first tentacle prototype has no taper ($R_T=1$), uniform tendon hole positions ($R_s=1$, or $b=c=15$~mm), and perpendicular crease lines $\alpha=\beta=90^{\circ}$. When the tendon is pulled, the tentacle primarily contracts linearly along its central axis and folds into an accordion configuration. The deformation is relatively uniform along the length and slightly concentrated near the base (Figure~\ref{fig:single}d$_i$, Movie S1).

\textbf{Tapering sequences the folding from the tentacle's tip}:
The second tentacle prototype introduces a tapered design with $R_T=0.5$, while keeping other design parameters unchanged. As a result, we observe a similar linear contraction deformation; however, the folding starts from the tentacle's tip and propagates towards the base --- this is a desirable deformation behavior for object grasping (Figure~\ref{fig:single}d$_{ii}$, Movie S1).

\textbf{Offsetting the tendon holes creates bending}:
In the third tentacle prototype, we change the tendon hole spacing ratio to $R_S=0.5$ (or $b=20$~mm $c=10$~mm). As a result, significant in-plane bending occurs. When $R_s=1$, the actuation tendon remains in the neutral axis of the folded origami structure; however, when $R_S \neq 1$ (or $b \neq c$), the tendon is offset away from the neutral axis, creating non-uniform folding between the neighboring crease lines (Figure~\ref{fig:single}d$_{iii}$, Movie S1). One can program such bending curvatures by tailoring the spacing ratio $R_S$.

\textbf{Re-orienting the crease lines adds twisting}:
Finally, in the last tentacle prototype, we change the crease angle $\beta$ from $90^{\circ}$ to $75^{\circ}$, while maintaining the tendon hole offset $R_S=0.5$. The non-perpendicular origami folds add twisting deformation that --- combined with the bending from offsetting tendon holes --- creates a three-dimensional, helical-like coiling deformation (Figure~\ref{fig:single}d$_{iv}$, Movie S1).

In summary, these four prototypes demonstrate how the tapering ratio $R_T$ primarily determines where folding initiates, the spacing ratio $R_S$ determines bending, and crease-line angles $\alpha$ $\beta$ determine twisting.  These design parameters are simple to tailor, allowing us to combine these deformation modes and create sophisticated 3D shapes strategically.

\subsection{Single Tentacle Deformation Mechanics}
To understand the underlying physics of the three-dimensional tentacle behaviors and \textit{quantitatively} map the origami design to deformation output, we derive a first-principle mechanics model. Since deformation of a single tentacle is driven by folding at each crease as the tendon that runs along the length of the tentacle is retracted, we
first determine the equilibrium configuration of individual folds under a specified level of tendon contraction by performing a series of force-balances along the tentacle.

Following established rigid-origami modeling approaches \cite{tachi2009simulation}, we describe the tentacle as a kinematic chain of rigid polygonal plates connected by revolute ``hinges'' along the crease lines (Figure~\ref{fig:single}c), with each crease hinge modeled as a linear torsional spring that produces a torque $\tau_i$ at the crease (see Methods). A force balance on a tentacle's elementary unit---spanning three hole locations $h_{i-1}$ to $h_{i+1}$ and containing two creases $c_i$ and $c_{i+1}$ (Figure~\ref{fig:single}c)---reveals equilibrium is achieved when the normal forces developed by creases $c_i$ and $c_{i+1}$ are of equal magnitude ($F_i-F_{i+1}=0$). The resulting distribution of folding angles is found to be a function of the hole-spacing pattern $d_i$ and local ribbon width $w_i$ via $\phi_i = kd_i/w_i$, where $k$ is a global scaling constant that enforces the overall tendon actuation length.  As a result, the model predicts that short-edge joints and wider segments yield less folding, while long-edge joints and skinnier segments fold more. With this model, the full set of folding angles $\{\phi_i | i=1..N\}$ can be found for a given level of tendon actuation, so we can reconstruct a forward-kinematic chain of rigid trapezoidal facets to simulate the overall tentacle shape (see Methods and SI for full details). Net 3D bending and twisting of the tentacle then emerges due to the tendon alternating between holes on opposite sides of the ribbon, which induces unequal folding across neighboring joints depending on the hole pattern.

Using the same design parameter as the experimental demonstration, the mechanics model reproduced the observed behaviors for all four tentacle prototypes, including taper-driven sequencing, in-plane bending, and the out-of-plane coiling (Figure~\ref{fig:single}e). This agreement supports treating tentacle response as a geometry-governed folding problem and enables rapid exploration of design effects without fabricating every variant.

\section{Stochastic and Emergent Tangling of Multiple Tentacles}

\begin{figure}[h!]
    \centering
    \vspace{-10pt}
    \includegraphics[width=0.99\textwidth]{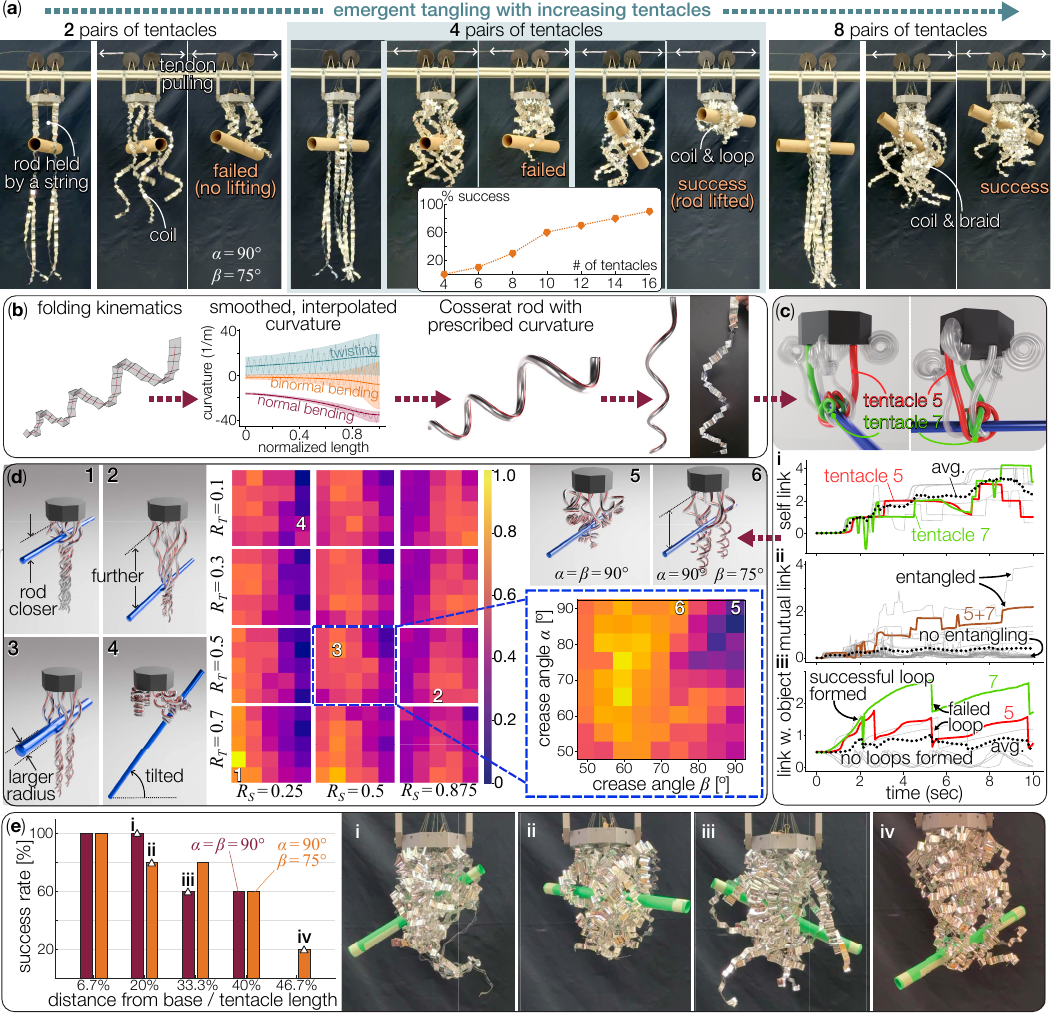}
    \vspace{-10pt}
    \caption{\textbf{Emergent tangling of multiple origami tentacles.}
    (a) Tangling behavior emerges as the number of tentacles increases. As a result, 2 pairs of tentacles cannot grip the target tube, 4 pairs have a moderate chance of success in grasping (here we show a failed and a successful test), and 8 pairs have close to 100\% of success (Movie S3). 
    (b) Modeling pipeline to connect the origami folding model to the dynamic Cosserat rod model. The predicted normal, binormal, and twisting curvature of the tentacle's equilibrium configuration under a given tendon actuation level is extracted from the origami-model (shaded region denotes curvature range), smoothed (solid lines), and used to update the reference curvature $\mathbf{\kappa}_0$ of the Cosserat rod tentacle, causing the rod to dynamically deform to match this new curvature. 
    (c) Link-based quantities of multi-tentacle entanglement. A representative case is shown with tentacle geometry  $\alpha=85^{\circ}$, $\beta=90^{\circ}$, $R_t=0.5$, and $R_s=0.5$. The coiling of two tentacles, \#5 (red) and \#7 (green), is highlighted, and the evolution of link-based quantities is shown.     
    Step changes in link quantities indicate crossings of a curve's tip boundary directors through the secondary curve, which discretely changes the computed number of crossings between the two curves. 
    (d) Parameter sweep of how tentacle design parameters impact grasping performance, defined as the sum of average object link and average mutual link (see Methods for definitions). Left: Visualization of grasping for different tentacle designs and obstacle setups. Center: Heatmap of grasping performance for different design parameters. Each sub-map corresponds to a unique spacing ratio ($R_s$) and taper ratio ($R_t$) with the x-axis and y-axis corresponding to the crease angles $\beta \in S$ and $\alpha \in S$, respectively where $S = \{ 50^{\circ}, 60^{\circ}, 70^{\circ}, 80^{\circ}, 90^{\circ}\}$. Right: Visualization of two tentacle configurations selected for experimental validation from refined parameter sweep heatmap (Movie S4).
    (e) Grasping experiments comparing two different origami designs, with representative successful captures shown in i–iv.
    }
    \label{fig:multi}
    \vspace{-10pt}
\end{figure}

While a single tentacle's deformation is determined by the underlying origami design, the collective behaviors of multiple tentacles arranged in parallel are instead governed by emergent and stochastic interactions. To understand this critical transition to emergent behaviors, we perform a comparative experiment (Figure~\ref{fig:multi}a).  In this experiment, we arrange tentacles in a circular array and mount them to a 3D-printed carrier. All tentacles have the identical coiling design as shown in Figure~\ref{fig:single}(d$_{iv}$) ($R_T=0.5,\, R_S=0.5,\, \alpha=90^\circ,\, \beta=75^\circ$). Moreover, tentacles were arranged in pairs. In each pair, one tentacle is designed to coil clockwise, and the other counterclockwise. This way, the neighboring tentacles in each pair coil in the opposite chirality, promoting inter-tentacle looping and braiding while maintaining an approximately symmetric overall layout. All tendons of these tentacles are connected to a pair of stepper motors to provide a simple and global pulling actuation. We then use a string to suspend a target object---a tube with 45~mm diameter, 280~mm length, and 15~gram weight---in between the tentacles and 150~mm from their base, corresponding to 20\% of the tentacle length, more experiment setup in Methods.

The experiments show that when the number of tentacles is small---e.g, 2 pairs---their coiling cannot create sufficient interaction between themselves and with the target, so no grasping is achieved.  However, as the number of tentacles increases, the \textit{possibility} of tangling between the tentacles \textit{and} with the target tube also increases. As a result, we observe an increasing chance of success in grasping, meaning that the origami tentacles can grasp and lift the tube above it original position.  When the number of tentacles is sufficiently high (e.g., 8 pairs), the grasping success rate converges to 100\% (see the inserts in Figure~\ref{fig:multi}a).

Critically, we observe two unique collective behaviors.  One is ``looping,'' in which the tentacles directly coil and wrap around the target tube.  The other is ``braiding,'' where the tentacles coil around each other, creating braid-like knots on the opposite side of the target object.  Then this knot hugs onto the object, achieving grasping and lifting. A close observation of the experiment results shows that braiding is a mechanically more robust approach toward robust grasping. These observations also suggest how these entanglement mechanisms scale with the number of tentacles. Adding more tentacles increases the admissible pathways for looping, braiding, mutual-link, and object-link, thereby increasing the probability of grasping success. This scaling benefit is most effective when the number of tentacles, carrier geometry, and tentacle spacing are coordinated as a design optimization.

Finally, it is worth noting that the entanglement and grasping are reversible upon tendon release. When the tendon actuation is reversed (e.g., reversing the servomotor), the tendon is unwound from the spool. Elastic energy stored in the folded Mylar material drives the tentacles back toward their relaxed configurations. This recovery allows the self-, mutual-, and object-link to disengage, enabling repeated grasping and release operations (Movie S3).

\subsection{Modeling the emergent behaviors from deterministic coiling}

To better understand such an emergence of tangling behavior, we extend our modeling framework to capture the dynamics of multiple tentacles as they mechanically interact with each other. To this end, we adopt a Cosserat rod formulation, wherein each tentacle is modeled as a single rod. A Cosserat rod is described by a center line position $\mathbf{\bar{x}}(s, t)$ and local rotation matrix $\mathbf{Q}(s, t)$ along its length $s \in [0,L(t)]$ \cite{Antman:2005,Gazzola:2018}. Each rod is capable of undergoing all modes of deformation---bend, twist, stretch, and shear---subject to any internal or external forces/torques (see Methods).  Internal forces/torques represent internal elastic strain forces and bending torques originating from the rod's elastic stiffness and deviation from its stress-free configuration; meanwhile, external forces/torques represent contact forces that may develop as the tentacle interacts with and entangles other tentacles. 

We integrate this Cosserat rod representation with our mechanics-based, single-tentacle origami model by matching the rod's stress-free curvature profile $\mathbf{\kappa_0}(s,t)$ to the curvature profile predicted by the single-tentacle origami model's force balance for a given fold pattern and tendon retraction level (Figure~\ref{fig:multi}b). This stress-free curvature matches the expected normal and bi-normal bending as well as twisting for a single tentacle actuated with no additional constraints, allowing us to define a rest configuration of the Cosserat rod based on our mechanics-based single tentacle model. Actively modulating the stress-free curvature of each rod then causes internal elastic torques and forces to develop within each tentacle proportional to how different the prescribed curvature is from the tentacles current curvature. 
These internal torques and forces, in turn, cause the tentacle to reconfigure itself as it seeks to equilibrate to its new (coiled) configuration.
Then, as tentacles coil, external contact forces develop as tentacles interact with and tangle with each other, thus allowing us to capture the dynamics of multiple tentacles interacting (see Methods for details). 
This assembly of rods is then numerically discretized and solved using \textit{Elastica} \cite{Gazzola:2018, Naughton:2021, PyElastica}, an open-source Cosserat-rod framework in Python whose utility has been demonstrated in a range of biophysical and engineering settings, including octopus arm manipulation \cite{Tekinalp:2024, Chang:2023}, snake locomotion \cite{Gazzola:2018, Zhang:2019, Zhang:2021}, spider web-inspired sensing \cite{Khairnar:2025, Khairnar:2026}, and soft robotic control \cite{Naughton:2021, Shih:2023, Naughton:2025}.

While our multi-tentacle model utilizes a validated framework of slender Cosserat rod dynamics and the stress-free intrinsic curvature of each tentacle is driven by our first principles mechanics model, the stochastic nature of multi-tentacle entanglement makes the precise force history of any particular pair of tentacles both irreproducible and largely irrelevant to the macroscopic grasp outcome. In an entangled grasp, the instantaneous contact forces and the failure load of any single grasp are themselves random variables that vary substantially between nominally identical trials. As such, attempting to resolve the detailed contact mechanics of every tentacle interaction has limited utility in understanding multi-tentacle grasping.

Instead, to gain insight we adopt a topological perspective rooted in knot theory to quantify the emergence of grasping from the entangling tentacles. 
Such topological quantities have been
long employed in biology to characterize the supercoiling morphology of DNA \cite{Bauer:1980, Fuller:1978} and has found increasing purchase in the analysis of slender structures \cite{Charles:2019, tekinalp2024topology, charles2025topological, becker2022active} as they provide a natural ordering for this class of system. In our case, they encode the structural reason a grasp holds (interlocking and topological locking) and remain regular across the stochastic ensemble even as individual forces fluctuate. As such, here we focus on analyzing the emergence of multi-tentacle grasping within this topological lens through the description of a tentacle \emph{link}, rather than exactly predicting specific tentacle forces and interactions. Link is the oriented crossing number (or Gauss linking integral) of two curves averaged over all projections. Practically, the link quantifies how much the two curves wind around each other. Based on this definition, we define three sub-quantities: \textit{Self-link} ($\link_s^i$) is the link of a tentacle's centerline with one of its edges (red auxiliary curve in Figure~\ref{fig:multi}b), \textit{Mutual-link} ($\link_m^{i,j}$) is the link of two tentacles' centerlines, and \textit{object-link} ($\link_o^i$) is the link of a tentacle's centerline with the longitudinal axis of a cylinder object being grasped (full definitions are provided in the Methods). 

To illustrate how these quantities relate to entangled grasping, in Figure~\ref{fig:multi}c we visualize a representative case where eight tentacles arranged in a configuration similar to the experiment in Figure~\ref{fig:multi}a are challenged to grasp a cylinder, with two tentacles (\#5 and \#7) highlighted to demonstrate how tentacles coil up and entangle with each other and the cylinder. As the tentacles coil up and entangle, the link-based quantities evolve (Figure~\ref{fig:multi}c${_{i-iii}}$). 

Self-link, which captures the overall level coiling of an individual tentacles, grows for all tentacles, with jumps in self-link corresponding to the addition or loss of a loop in the tentacles configuration (Figure~\ref{fig:multi}ci). Mutual link then quantifies the level of entanglement between two tentacles. Tentacles that coil up (self-link increases) but are topologically separated (no entanglement) have low to no mutual-link. Indeed, the majority of tentacle pairs exhibit this relationship (Figure~\ref{fig:multi}c${_{ii}}$). However, tentacles that do entangle with each other, such as tentacles \#5 and \#7, exhibit increasing mutual-link. Finally, object-link quantifies how much a tentacle has wrapped around the object. Tentacles that do not interact with the object have little to no object-link (Figure~\ref{fig:multi}c${_{iii}}$), while tentacles that entangle with the object increase their object-link as more loops are created. As such loops begin to form, object-link steadily increases. If, however, the loop being formed by the tentacle curls up before wrapping around the cylinder, the object-link drops accordingly. This can lead to the ``oscillation'' of object-link as coils are formed but ultimately do not wrap around the object.  Therefore, the average object-link serves as a better proxy of the number of loops a tentacle has formed around the object. For example, after tendon pulling, tentacle \#5 has wrapped roughly 1 turn and tentacle \#7 has wrapped 2 turns around the objects, which are reflected in their respective object-link values exhibiting ranges of $\sim$0.5--1.0 and $\sim$1.5--2.5.

\subsection{Optimizing entangled grasping performance}

With our link-based quantities, we can examine how different tentacle designs relate to grasping performance in air and under gravity. To do so, we define a grasping performance function consisting of the summation of the average mutual-link and average object-link quantities over all tentacles (see Methods for full definition). Self-link is not used as it does not directly contribute to entanglement.

Using this grasping performance function, we perform a parameter sweep across four tentacle design parameters: crease angles $\alpha$ and $\beta$, taper ratio $R_T$, and spacing ratio $R_S$ to generate heatmaps of grasping performance (Figure~\ref{fig:multi}d, Movie S4). 
For each case, we perform five grasping trials with varying object locations, orientations, and radii (Figure~\ref{fig:multi}d${_{3-6}}$) to assess grasping robustness and generality. 
The spacing ratio $R_S$ and taper ratio $R_T$ have limited impact on grasping performance, although performance does somewhat improve as the spacing ratio deviates from unity, which increases bending, and as the taper ratio increases, which yields a more flexible distal segment of the tentacles, so that the tentacles bend earlier at their tip and interact more readily with neighboring tentacles.  Crease angles---$\beta$ in particular---exhibit a much stronger impact on grasping performance, with better grasping performance generally achieved at either a lower $\alpha$ or $\beta$ angle. Representative grasping cases of different tentacle configurations are provided in Figure~\ref{fig:multi}d${_{3-6}}$ and in the Movie S4. 

Overall, two tentacle morphologies are found to exhibit successful grasping behavior. The first, exhibited in Figure~\ref{fig:multi}d${_{3-5}}$, achieves grasping by wrapping the distal ends of tentacles around each other into unified bundles (i.e., braiding), effectively locking the free-end boundary condition and securing the object between the base and the bundles. The second, exhibited in Figure~\ref{fig:multi}d${_{6}}$, directly wraps tentacles around the object without extensive entanglement with neighboring tentacles (i.e., looping). Both morphologies successfully grasp the object, illustrating how both mutual-link and object-link contribute to the emergence of entangled grasping by the tentacles (see SI for additional discussions).

Based on this initial parameter sweep, we next perform a refined parameter sweep across crease angles for tentacles with a moderate taper ratio of $R_T$=0.5 and a spacing ratio of $R_S$=0.5 (Figure~\ref{fig:multi}d, lower left), as this configuration exhibits robust grasping performance across crease angles. To better identify how stochastic variability in tentacle entanglement impacts grasping, the refined parameter sweep considers a single baseline object position over five different trials with random perturbations injected into the tentacle activation signal. From this refined map, we select two representative designs that are compatible with our fabrication method for further experimental tests. The first design (Figure~\ref{fig:multi}d${_{1}}$) has crease angles of $\alpha$=$\beta$=$90^\circ$ while the second design (Figure~\ref{fig:multi}d${_{2}}$) has crease angles of $\alpha$=$90^\circ$ and $\beta$=$75^\circ$. The second design is predicted to exhibit better grasping performance, which we next experimentally validate (Movie S4).

\subsection{Experimental Validation}
To validate the performance trends from the Cosserat rod simulations, we fabricated and compared two representative multi-tentacle grippers exhibiting distinct single-tentacle deformations. The first design is bending-dominant with $\alpha=\beta=90^\circ$; These tentacles would still coil due to gravity (see SI Section 4), but they would not generate as much mutual-link or object-link, so they serve as the lower-performing reference. The second design involves significant twisting with $\alpha=90^\circ$ and $\beta=75^\circ$, so it should exhibit improved entanglement performance according to simulation results in Figure \ref{fig:multi}d. Notice that the optimal design $(\alpha,\beta)=(75^{\circ},60^{\circ})$ was not tested because its more aggressive fold geometry accelerated wear along the crease lines under repeated actuation.

We quantify the grasping performance of these two designs using the success grasping rate over 10 repeated trials. Each trial involves 8 tentacle pairs, and the rod is suspended at a prescribed distance from the tentacles' base, ranging from 6.7\%–46.7\% of the tentacle length. When the rod is close to the base (6.7\%), both designs achieve consistent grasping, indicating that a relatively short reach allows the tentacles to entangle sufficiently to envelop and lift the rod. As the rod moves away from the base, the success rates decline, and the two designs diverge in performance. More specifically, the bending-dominant $\alpha=\beta=90^\circ$ gripper starts to show grasping performance dropping early when the rod distance is only 20\%, and it completely fails when the rod distance is close to 50\% (Figure~\ref{fig:multi}e).  In contrast, the twisting-dominant gripper maintains a higher success rate and can still achieve occasional grasps at the highest rod distance.  This suggests that twist-induced, additional coiling increases mutual and object-link, promoting mechanical ``locking'' through entanglement. Representative successful grasps at selected rod distances illustrate this transition from near-base capture to more geometry-dependent entanglement as the target is placed farther from the carrier (Figure~\ref{fig:multi}e${_{i-iv}}$).

\section{Robust Object Grasping in Diverse Operation Conditions}
\label{subsubsec:grasping_testing}

With an understanding of deterministic tentacle deformation and their stochastic tangling, we then study the robust grasping performance of a multi-tentacle system under different operating conditions, including in air under gravity, underwater with buoyancy force, and in space without gravity.

\subsection{Grasping In Air}

\begin{figure}[t!]
    \centering
    \includegraphics[width=\textwidth]{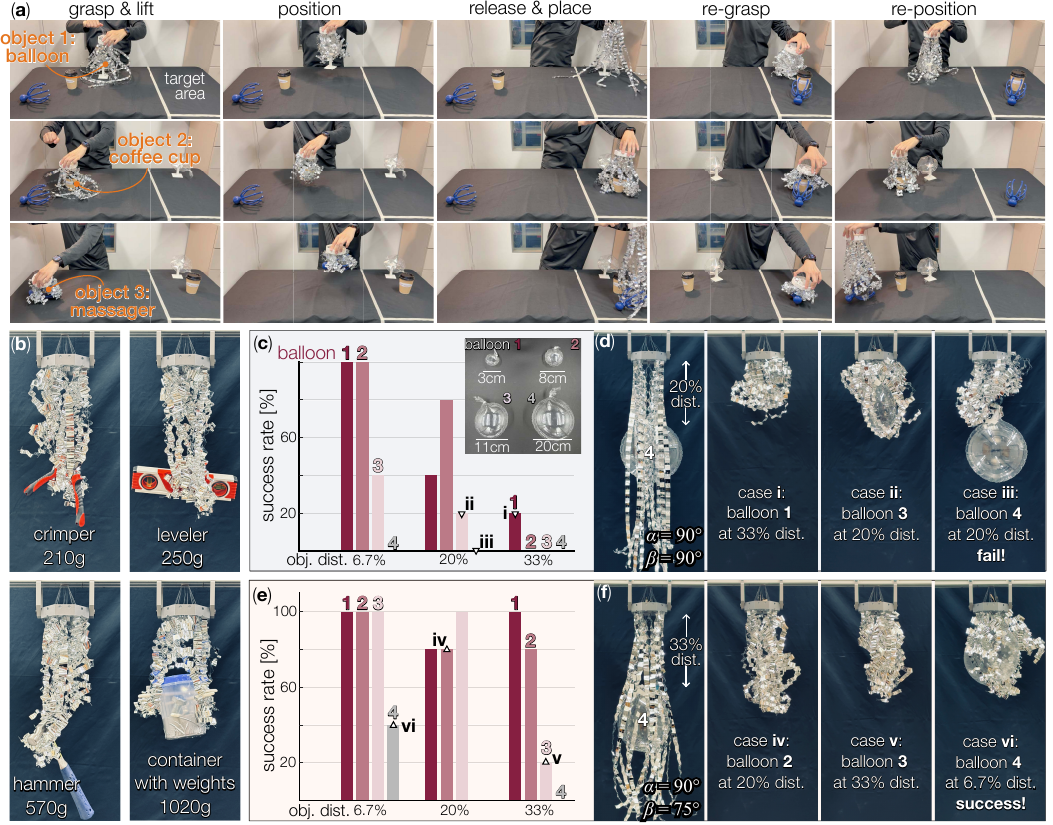}
    \vspace{-20pt}
    \caption{\textbf{Robust object grasping in air (under gravity).} (a) Realistic table-top manipulation demonstration showing grasping, positioning, release, re-grasping, and re-positioning of objects with different shapes and surface properties (Movie S5). (b) Heavy object grasping with sixteen, $\alpha=90^\circ$ $\beta=75^\circ$, tentacles to evaluate their collective, maximum payload capacity. (c,d) Balloon-grasping tests with bending-dominant $\beta=90^\circ$ tentacles, with a few successful examples. (e,f) Balloon-grasping tests with twisting $\beta=75^\circ$ tentacles (Movie S6).}
    \label{fig:air}
\end{figure}

We evaluate the robustness of the entangling tentacle grasping through two sets of experiments. First, we conduct a realistic demonstration in which the manually driven tentacle gripper grasps, positions, releases, and re-grasps complex objects from the table top.  Next, we conduct a series of more controlled tests to examine the weight and volume capacity, repeatability, and disturbance rejection of tentacle grasping. Note that all tests are completed by 8 pairs of $\alpha=90^\circ$, $\beta=75^\circ$ tentacles.

\underline{Realistic Tabletop Grasping Demonstration}: We place three objects on the tabletop: a small, smooth spherical object (a balloon), a complex and irregular object (a head massager), and a fragile object (a paper coffee cup filled with approximately half a cup of coffee). The tentacle gripper successfully approaches these objects from a distance, grasps them, repositions them, releases them, and subsequently re-grasps them, all in a continuous operation without excessive squeezing, damage, or coffee spillage (Figure~\ref{fig:air}a and Movie S5). This demonstration highlights the real-world practicality of the tentacle grasping.

\underline{Weight Capacity}: In this test, we use the same 8 pairs of tentacles to grasp objects with increasing weight, including a crimper (210 g), a leveler (250 g), a hammer (570 g), and a weighted container with spokes (1020 g) (Fig.\ref{fig:air}b). Each tentacle and its tendon weighs only $\sim2$ g, giving a total tentacle mass of $\sim32$ g.  Therefore, the entangling tentacles can lift objects to 31 times their own weight, validating their heavy-lifting potential.

\underline{Volume Capacity}: We use the same tentacle setup to grasp large plastic balloons, which present a smooth, low-friction surface that is difficult to secure with conventional grippers. Four balloon sizes are tested (3~cm, 8~cm, 11~cm, and 20~cm diameter). The balloons are suspended at four different positions at the beginning of each trial, so the initial distance between the tentacle's base and the balloon's top is 6.7\%, 20\%, and 33.3\% of the tentacle length. Five trials are tested for each setup. The bending-dominated $\alpha=\beta=90^\circ$ design achieved a high success rate for small balloons at the shortest distance (6.7\%), but it failed consistently at larger balloon sizes and distances (Figure~\ref{fig:air}c,d). In contrast, the twisting $\alpha=90^\circ$, $\beta=75^\circ$ design showed substantially improved performance across balloon sizes and distance---including successful grasps of the largest balloon and more consistent success rates for smaller balloons (Figure~\ref{fig:air}e,f). Grasping video footage showed a qualitative transition from simple coiling from the $\beta=90^\circ$ design to inter-tentacle braiding and knotting from the $\beta=75^\circ$ design (Movie S6). Such a deformation-mode transition improves grasping performance, which is consistent with the entanglement mechanism identified in the simulations.

\underline{Repeatability:} We conduct a 211-cycle repeatability test (Movie S7 and SI Section 8). In the first cycle, we use the same tentacle gripper to grasp and release a tree-like object. Then, we repeatedly coil and release the tentacles without a target object from Cycles 2 to 100. Cycle 101 is a second grasp-and-release operation, followed by another sequence of coiling and releasing during Cycles 102--200. Finally, during Cycles 201--211, the tentacles grasp and release the same tree-like object 11 times. Across all cycles, we observe no noticeable degradation in grasping performance. We attribute this durability primarily to the distributed nature of the coiling deformation and the conditioning step incorporated during fabrication.

\underline{Disturbance Rejection:} To further evaluate grasping stability, we performed a disturbance test (Movie S8 and SI Section 9).  In this test, the tentacles grasp the tree-like object and maintain the grasp continuously for 60 seconds. During this period, we repeatedly struck the object with a metal bar, 15 impacts from each of four different directions. Critically, experiment footage reveals that some object links can slip or fail under disturbance, but the overall grasp remains stable. The multiple links with object shown in Figure \ref{fig:multi}(c) inherently create redundancy, allowing the remaining links to compensate for local failures and maintain a stable global grasp under external disturbances.

\subsection{Grasping  Under Water} 

\begin{figure}[t!]
    \centering
    \includegraphics[scale=1.0]{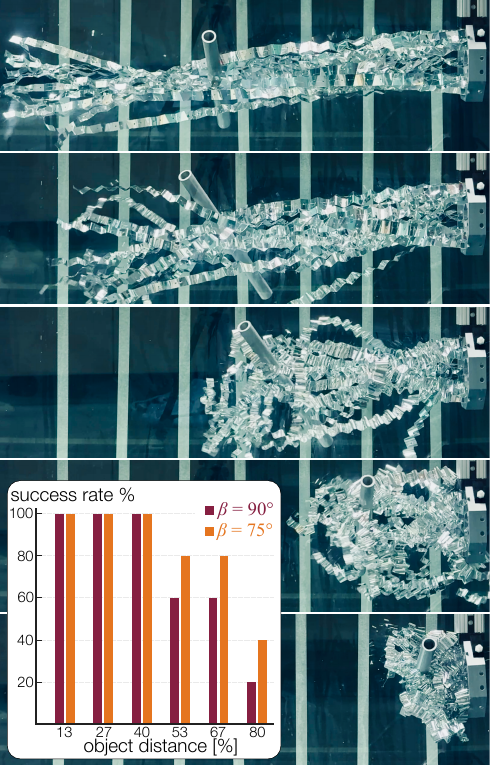}
    \caption{\textbf{Object grasping underwater.} The snapshot photos show the sequence of a representative grasp, while the insert figure summarizes the grasping success rate of the $\beta=90^\circ$ and $\beta=75^\circ$ tentacles (Movie S9).}
    \label{fig:water}
\end{figure}

The Mylar plastic origami tentacles have approximately neutral buoyancy, and the tendon actuation works well underwater.  Therefore, we evaluate multi-tentacle grasping performance in the underwater environment.  Both tentacle designs ($\beta=90^\circ$ and $\beta=75^\circ$) are tested using the same 16-tentacle setup. However, everything is placed horizontally rather than vertically: The carrier was rigidly mounted to a frame, facing sideways, while the target rod is suspended by a string at a prescribed horizontal distance from the carrier. For each grasping trial, we first manually suspend the tentacles straight and horizontal in the water, with the rod centered underneath. Then, the tendons are pulled manually to execute the grasp (Figure~ \ref{fig:water}).  Six rod distances were tested (13.3\%–80\%), with five repeated trials per distance and tentacle design.

Surprisingly, the entangling tentacles demonstrate a substantially larger grasping reach than their in-air performances. Both designs successfully captured the rod at a distance up to 80\% (insert of Figure~\ref{fig:water}a). Grasping performance is consistently high at short-to-moderate distances: both designs achieved a 100\% success rate up to 40\% object distance. At larger distances, success rates of the two tentacle designs diverge: Bending-dominant $\beta=90^\circ$ tentacles show decreasing performance with 60\% and 20\% success rates when the rod is 53.3\% and 66.7\% away, respectively. Meanwhile, the twisting $\beta=75^\circ$ tentacle retained a more robust grasping, with 80\% and 40\% success rate at the same rod distances (Movie S7). Overall, the tentacles can entangle more easily underwater, and the floating rod can be moved with a smaller force --- these factors explain the improved grasping performance.

\subsection{Deployment and Grasping in Zero Gravity}

Space is another environment where the entangling tentacles could work well. The origami tentacles are lightweight and require simple actuation; therefore, they could be rolled up into a small volume---i.e., with a high packing ratio---into compact payload packages in a host satellite for rocket launch. Once in space, the host satellite can deploy the tentacles via springs or spin, and then grasp objects (e.g., space debris cleaning, in-orbit manufacturing assembly, or satellite servicing).  The emergent nature of the entangling grasping means that the origami tentacles can handle objects with complex shapes and orbital dynamics without extensive adapters or actuators.

\begin{figure}[t!]
    \centering
    \includegraphics[width=\textwidth]{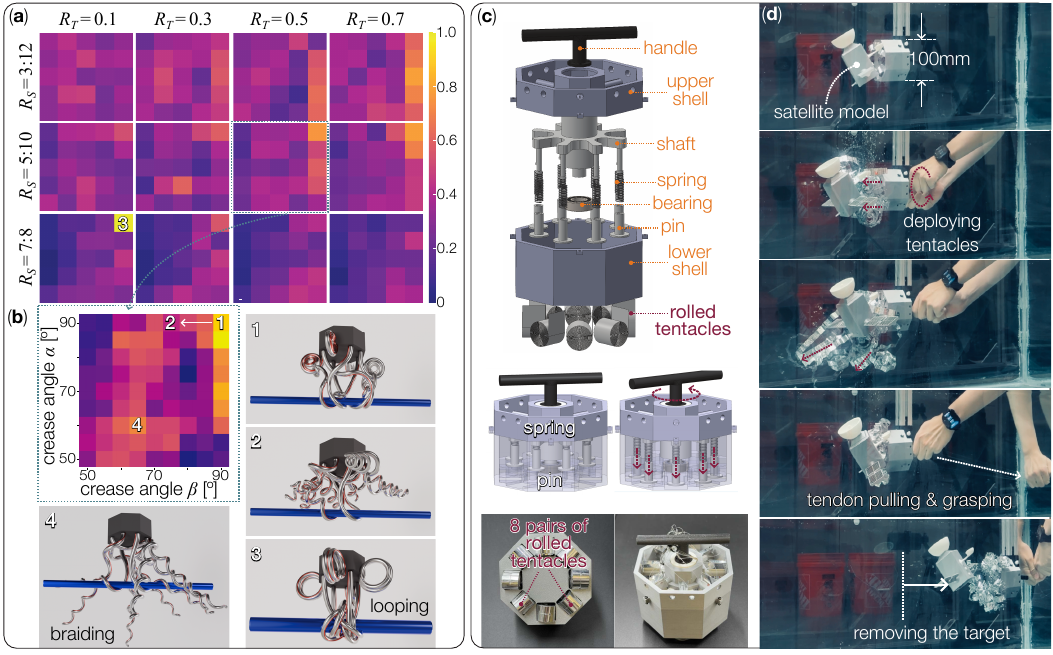}
    \caption{\textbf{Simulation-guided design sweep without gravity and simulated underwater demonstration of in-space capture.} (a) Coarse parameter sweep showing the entanglement score over folding angles ($\alpha$, $\beta$) for multiple spacing ratios and taper ratios. (b) Refined map at the selected spacing/taper condition, with representative configurations illustrating two dominant interaction modes (looping and braiding) at marked design points (Movie S10). (c) Deployment mechanism design. (d) Underwater demonstration sequence mimicking a flyby-and-capture task: locate target, deploy tentacles, initiate entanglement, pull tendons to secure grasping, and translate the captured target (Movie S11).}
    \label{fig:space}
\end{figure}

Therefore, we assess the entangling origami tentacles' grasping performance in a zero-gravity environment using Cosserat rod simulations and demonstrate a possible deployment setup in a simulated underwater test. To do so, we adopt the same optimization framework of Figure~\ref{fig:multi}, but without gravity in the simulation, to perform a parameter sweep of how different tentacle designs relate to grasping performance in a zero-G environment (Figure~\ref{fig:space}a). We find entanglement in zero-g is substantially different from that in one-g. Previously, gravity pulled tentacles downward, providing a consistent orientation that helped create large multi-tentacle braids for grasping. In contrast, without such gravitational regularization, helically-coiling tentacles tend to deflect laterally, reducing their inter-tentacle interactions. While ``braiding'' entanglement does occur at smaller crease angles ($\alpha=60^{\circ}$; $\beta=70^{\circ}$) as seen in Figure~\ref{fig:space}b${_{4}}$, they consist of only a few tentacles at a time, rather than all tentacles forming a single large braid as in the gravity case (Figure~\ref{fig:multi}d). A second regime of successful grasping corresponds to a ``looping'' behavior that emerges at large crease angles ($\alpha=90^{\circ}$; $\beta=90^{\circ}$) as demonstrated in Figure~\ref{fig:space}b${_{1, 3}}$. Here, rather than forming helical coils, tentacles form large planar loops that allow them to directly grab the object by wrapping around it (Movie S10). 

To simulate a deploy and grasping operation in a space-like environment, we performed an underwater demonstration that mimics a satellite flyby and capture. When underwater, buoyancy cancels gravity, allowing the tentacles to behave somewhat as if they are in space. We rolled 8 pairs of $\beta=75^\circ$ origami tentacles into compact bundles and fit them into a custom-designed, 3D-printed casing, which has a similar size to a 1U CubeSat (Figure~\ref{fig:space}c). Once in water, deployment was triggered by a simple mechanical release mechanism: A central shaft can rotate and align a set of eight push pins with their exit paths, then coil springs on this shaft push these pins outward, ejecting the rolled tentacle bundles to complete a rapid deployment. The deployed tentacles would then unroll and gradually enclose the target object: a 3D-printed satellite model in this case. Then, we manually pull the tendons to coil and entangle the tentacles, achieving grasping.  The grasping force is sufficient to move the satellite model into a different location (Figure~\ref{fig:space}d, Movie S11). It is worth emphasizing that in this simulated test, the drag force from water significantly hindered the deployment reach of origami tentacles.  If they were in space, we expect them to deploy much further, meaning a better grasping performance.

\section{Discussion and Conclusion}

By synergizing local, deterministic deformation programming and global, stochastic entangling, this study presents an origami tentacle gripper that can grasp complex-shaped objects under different working environments, while preserving simplicity in design and actuation. More specifically, we design an origami tentacle featuring alternating holes and crease lines, and actuate it by pulling a tendon that weaves through these holes and attaches to the tentacle’s tip. By choosing the tendon hole spacing, crease line orientation, and overall tapering, one can determine the deformation pathway of the tentacle and achieve 1D shrinking, 2D bending, and 3D coiling. As multiple coiling tentacles are placed in proximity, collective behaviors emerge: they entangle and braid themselves into a knot to enclose an object or loop around it, thus securing a robust grasp. 

\medskip
Such results support a grasping principle centered on contact-driven self-organization: robust capture can be achieved by geometry-programmed coiling and emergent entanglement rather than precise sensing and control. To quantitatively correlate origami design, deterministic deformation, and stochastic entangling, we formulate a simulation model that integrates origami mechanics and Cosserat rod simulation. These simulation results guide us to the fundamental physics: the Mutual-link and object-link are the key performance metrics.  Simulation also provides optimized tentacle designs, which allowed us to test and demonstrate robust object grasping in air (under gravity), underwater, and even in space without gravity. 

\medskip
Several practical limitations motivate future work. Aggressive fold geometries accelerate material wear at the crease after repeated folding/unfolding, suggesting the need for improved materials \cite{guo2025light, li2026leaf}, local reinforcement, or replaceable crease elements. The current tests mostly focus on canonical targets, and extending to grasp more irregular objects and especially moving targets might require more tightly integrated actuation, novel deployment strategies, and perhaps onboard sensing. Scaling up tentacle length to grasp larger objects is another open challenge that requires us to further probe into the emergent behaviors. Regardless, the entangling origami tentacle gripper is a promising mechanism for scalable and robust object manipulation in terrestrial, underwater, and space-relevant environments.

\section*{Methods}

\subsection*{Origami tentacle fabrication}

We fabricate the tentacles by cutting Mylar (PET) film sheets using a \textbf{Graphtec Cutting Pro FCX4000-60}. The cut pattern was generated in \textbf{Graphtec Pro Studio} and exported as a vector file for cutting. The cutting speed is 7 cm/s at a force of 35~N, with a blade length of 0.8~mm. Once removed from the cutter, each tentacle underwent an additional conditioning step to improve repeatability. Specifically, we manually fold the ribbon along the programmed crease lines to the maximum deformation, then place it in a 3D-printed mold for \textbf{8 hours} to release residual stress in the creases. Because of their different geometries, the $\beta=90^\circ$ and $\beta=75^\circ$ tentacles are conditioned in different mold orientations to match their preferred folding and coiling directions.

\subsection*{Modeling origami folding kinematics of a single tentacle}

To connect folding patterns with tentacle kinematics, we formulate a numerical solver based on established rigid-origami modeling approaches \cite{tachi2009simulation}. The solver is implemented in Python and made freely available online (see Data Availability statement). A tentacle of length $L$ is modeled as a kinematic chain of $N+1$ rigid polygonal plates connected by $N$ revolute ``hinges'' created by the creases between plates. Each crease $c_i$ is modeled as a linear torsional spring with bending torque $\tau_i=2B_i\phi_i$, where $\phi_i$ is the folding angle defined as one half of the outer angle of the $i$-th joint. The bending stiffness $B_i=EI_i$ depends on the elastic modulus $E$ of the tentacle material and the local second moment of area $I_i=t^3w_i/12$, where $t$ is the thickness and $w_i$ is the width of the tentacle at the crease location. Tendon routing between holes is the straight line distance $\ell_i=2d_i\cos(\phi_i)$ between adjacent holes $h_i$ and $h_{i+1}$ where $d_i$ is the distance between crease $c_i$ and hole $h_i$.

To determine the tentacles' configuration given a level of tendon activation, we define elementary units consisting of three holes and two creases. The unit spans from hole location $h_{i-1}$ on one side to $h_{i+1}$ on the other and contains creases $c_i$ and $c_{i+1}$. We then perform a force balance around hole location $h_i$, with the torque from each crease developing a normal force $F_i = \tau_i/2d_i$ at hole location $h_i$. Equilibrium is achieved when the magnitude of the two forces balances such that $F_i-F_{i+1}=0$. Note that here the torque developed by each crease is split between adjacent elementary units, leading to a factor of two. This, in turn, can be reduced to  
$w_i\phi_i/d_i - w_{i+1}\phi_{i+1}/d_{i+1}=0$, which reveals equilibrium is achieved for all $\phi_i=kd_i/w_i$, where $k$ is a scaling constant related to the shortened tendon length $L_t(\gamma)=(1-\gamma)L$ given a tendon actuation parameter $\gamma$. Our experimental tentacle setup is driven by position-based control of the tendon contraction level, thus our model considers equilibrium under a prescribed tendon length and not under a tendon actuation force, though extension of our model to such a case is straightforward. The summed length of individual tendon length between holes is equal to the overall actuated tendon length $\sum_{i=1}^N \ell_i = L_t(\gamma)$, which yields 
\begin{equation}
\sum_{i=1}^N 2d_i\cos\left(\frac{kd_i}{w_i}\right)=(1-\gamma)L 
\quad \text{for} \,\,\,\, \phi_i=\frac{kd_i}{w_i}\in{\left[0,\frac{\pi}{2}\right]}.
\label{eqn:solver}
\end{equation}
We note that $k$ is a global constant since each crease appears in two elementary units, thus chaining them together and requiring the same $k$ for each elementary unit force balance.

Under this formulation, the original multi-variable configuration problem reduces to finding the global scaling term $k$ that satisfies the global length constraint. Once $k$ is determined, the full set of folding angles $\{\phi_i\}$ follows directly from $\phi_i=kd_i/w_i$. Because the left-hand side is monotonic in $k$ over the prescribed interval $\phi_i\in(0,\pi/2)$, the solution can be efficiently obtained using a one-dimensional bisection root-finding procedure. If one or more folding angles $\phi_i$ reaches $\pi/2$ under large tendon displacements, we incorporate an extended iterative approach to handle this case which is described in the SI.  

\subsection*{Cosserat rod modeling of multiple tentacles}

To model the dynamic entanglement of tentacles, we consider each tentacle as a Cosserat rod, which are slender, one-dimensional elastic structures that can undergo all modes of deformation: bending, twisting, stretching, and shearing. This representation entails several key advantages. Multiple rods can be assembled together to capture the interactions of tentacles are they entangle and each rod's 1D representation can be connected to the prior origami kinematic model through modulation of its passive reference curvature.  Their numerical implementation is also computationally efficient as they accurately capture large 3D deformations through a one-dimensional representation, alleviating time-consuming remeshing steps and compute costs of 3D elasticity. 

\textbf{Dynamics of a single Cosserat rod.}
An individual Cosserat rod is described by its center line position $\mathbf{\bar{x}}(s,t) \in \mathbb{R}^{3}$ and an oriented reference frame of orthonormal directors $\mathbf{Q}(s,t) \in \mathbb{R}^{3 \times 3} =\left[\mathbf{\bar{d}}_{1}, \mathbf{\bar{d}}_{2}, \mathbf{\bar{d}}_{3}\right]^{-1}$ along its length $s \in [0, L(t)]$, where $L(t)$ is the current length, for all time $t \in \mathbb{R} \ge 0$. 
The global lab frame ($\mathbf{\bar{v}}$) and local frame ($\mathbf{{v}}$) are then connected via $\mathbf{v} = \mathbf{Q}\mathbf{\bar{v}}$. For an unsheared rod, $\mathbf{\bar{d}}_{3}$ is parallel to the rod's local tangent ($\partial_s \mathbf{\bar{x}} = \mathbf{\bar{x}}_{s}$), and $\mathbf{\bar{d}}_{1}$ (normal) and $\mathbf{\bar{d}}_{2}$ (binormal) span the rod's cross-section. In the local frame, the components of the curvature vector $\bm{\kappa}(s,t) \in \mathbb{R}^{3} = [\kappa_1,~\kappa_2,~\kappa_3]$ relate to bending (${\kappa}_{1}$ and ${\kappa}_{2}$) and twisting (${\kappa}_{3}$) of the rod. The linear velocity of the center-line is $\mathbf{\bar{v}}(s,t) \in \mathbb{R}^{3}=\partial_t \mathbf{\bar{x}}$, the angular velocity vector $\bm{\bar{\omega}}(s,t) \in \mathbb{R}^{3}$ is determined through the relation $\partial_t \mathbf{\bar{d}}_{j} = \bm{\bar{\omega}} \times \mathbf{\bar{d}}_{j}$, and the second area moment of inertia $\mathbf{I}(s,t)\in \mathbb{R}^{3 \times 3}$, cross-sectional area ${A}(s,t)\in \mathbb{R}$, and density $\rho(s)\in \mathbb{R}$ are defined based on the rod's material properties.
The dynamics of a Cosserat rod are then described through the conservation of linear and angular momentum \cite{Gazzola:2018} 
\begin{equation} 
    \partial_{t}^{2} \left( \rho A \mathbf{\bar{x}} \right)= \partial_s \left(\mathbf{Q}^{T} \mathbf{n}\right) + \mathbf{\bar{f}}
    \label{eqn:linear_momentum}
\end{equation}
\begin{equation}\label{eqn:angular_momentum}
    \partial_t \left( \rho \mathbf{{I}} \bm{\omega} \right) = \partial_s \bm{\tau} + \bm{\kappa} \times \bm{\tau} + \left(\mathbf{Q} \mathbf{\bar{x}}_s \times \mathbf{n}\right)
     + \left(\rho \mathbf{I} \bm{\omega}\right) \times \bm{\omega} +  \mathbf{Q}\mathbf{\bar{c}}
\end{equation}
where Eq.~\ref{eqn:linear_momentum} (lab frame) and  Eq.~\ref{eqn:angular_momentum} (local frame) represent the change of linear and angular momentum at every cross-section, respectively, $\mathbf{n}(s,t)\in \mathbb{R}^{3}$ and $\bm{\tau}(s,t)\in \mathbb{R}^{3}$ are internal forces and couples, respectively, developed due to elastic deformations while $\mathbf{\bar{f}}(s,t)\in \mathbb{R}^{3}$ and $\mathbf{\bar{c}}(s,t)\in \mathbb{R}^{3}$ capture external forces and couples applied to the arm, respectively. To numerically solve the above continuous representation, rods are discretized into $\left(n_{\text{elem}}+1\right)$ nodes connected by $n_{\text{elem}}$ cylindrical elements, and the rods dynamics are computed by integrating the discretized set of equations along with appropriate boundary conditions. To do this, we used PyElastica \cite{PyElastica}, which is an open-source, Python-based implementation of this numerical scheme.

\textbf{Single tentacle actuation.}
Internal elastic couples in the rod are modeled using a linear elastic relationship $EI(\mathbf{\kappa}(t,s) - \mathbf{\kappa}_0(t,s))$ where $\kappa_0(t,s)$ is a reference rest curvature of the rod and captures both bending and twisting. When a tentacle's tendon is shortened, it causes the tentacle to assume a new equilibrium shape. To actuate our Cosserat rod tentacle, the rod's reference curvature $\mathbf{\kappa}_0(t,s)$ is dynamically changed to match the tendon curvature predicted by the origami-based kinematic model above. This then causes the rod to dynamically reconfigure itself to reach this new reference curvature. 

Specifically, the reference curvature $\mathbf{\kappa}_0$ is set to match the oriented tendon curve determined by the origami model (not the zig-zag orientation of the folded ribbon). This tendon curve is represented by a oriented piecewise curve of straight segments that connect hole locations along the tentacle, with each hole location having a local orientation matrix $Q_i$ that capture the normal, binormal, and tangent directors of the hole. The curvature of the $i$-th tendon segment is numerically evaluated as:
\begin{equation}
    \kappa_i=\frac{\log(Q_{i+1}Q_i^T)}{D_i},
\end{equation}
where $\log(\cdot)$ is a matrix logarithm operator $\mathbb{R}^{3\times3} \rightarrow \mathbb{R}^{3}$, $Q_i$ is the rotation matrix of $i$-th joint, and $D_i=(\ell_i+\ell_{i+1})/2$ is the average length of tendon around $i$-th joint.
Because curvature in the origami-based model is defined on a discretized structure of 49 rigid origami faces, the resulting curvature profile is piecewise. To obtain a smooth target curvature compatible with the Cosserat rod discretization, the curvature is smoothed using a one-dimensional Gaussian filter and interpolated into $N=100$ elements to match the Cosserat rod discretization.

Each tentacle is modeled as a circular cross-section tapered rod. Geometric parameters are set to match those of the experimentally fabricated tentacles such as the radius at the base $r_{b}$=2 cm and overall length $L$=75 cm. To compensate for differences in cross-sectional area and second moment of inertia between the physical ribbon and the circular rod approximation, the rods Young's modulus and density are reduced to $E$=100 kPa and $\rho$=100 kg/m\textsuperscript{3}, respectively. 

\textbf{Interactions of multiple arms and objects.}
Combining multiple rods allows the simulation of a multi-tentacle gripper.
The simulated gripper consists of eight tentacles arranged at the vertices of a regular octagon, suspended vertically with each tentacle's base fixed. At rest, the surface-normal direction of tentacles alternates between pointing towards the center of the octagon and pointing radially outwards so as to combine different winding handednesses. Additionally, tentacles are slightly tilted toward the central axis to increase contact likelihood. A fixed rigid cylindrical rod is initialized between the tentacles at a designated distance and orientation, which the multi-tentacle gripper is challenged to grasp.

The gripper is actuated at a constant rate corresponding to $3\%$ of the total tendon length per second.  While Cosserat rod's are capable of incorporating shortening and lengthening strain effects\cite{gazzola2018forward,Tekinalp:2024}, the large length change that tentacles undergo due to tendon shortening is difficult to fully incorporate using traditional strain/dilatation effects. To account for this, we model coiling behavior under a constant-length assumption which we then map to the shortened tentacle configuration by dynamically increasing the spacing between the base of tentacles such that their distance from each other is proportional to that of the shortened tentacle. 

As tentacles coil, they begin to entangle with each other.  To model the contact forces that develop between tentacles, we consider the forces that arise from contact between two Cosserat rod elements $R_i$ for $i\in\{1,2\}$ (or similarly, between a Cosserat rod element and the solid cylinder). We consider contact to occur when $\epsilon_{i} \ge 0$ for $\epsilon_{i}=r_{i}+r_{j}-|\bm{\delta}^{c}_{i}|$ where $\bm{\delta}^{c}_{i}=\bar{\mathbf{x}}^{c}_{j}-\bar{\mathbf{x}}^{c}_{i}$ is the distance vector between the center location of the elements with radius $r_i$ for $i\in\{1,2\}$ and $j\in\{2,1\}$.
We model contact between these two elements as a repulsive force $\bar{\mathbf{F}}_{i}$ on element $R_i$ based on the Hertzian contact model 
$\bar{\mathbf{F}}_{i}= -H(\epsilon_{i}) k_{c} \epsilon_{i}^{3/2} \hat{\bm{\delta}}^{c}_{i} \,\,\, \forall \,\,\, i\in\{1,2\}$
where $H(\cdot)$ is a Heaviside function that ensures a force is produced only in case of contact $\left(\epsilon_{i} \ge 0 \right)$, $\hat{\bm{\delta}}^{c}_{i}=\bm{\delta}^{c}_{i}/|\bm{\delta}^{c}_{i}|$, and $k_c$ is the contact stiffness coefficient which is a function of the material and geometric properties of the rods \cite{Hertz:1882}. This (soft) approach allows some amount of penetration to occur during tentacle entanglement; however, previous quantification of the degree of penetration shows it to be small \cite{Tekinalp:2024}.  When relevant, the effect of gravity is molded as a global body force applied to all tentacles. 

To capture the stochastic nature of multi-body interactions, each tentacle is assigned a random actuation delay $t_{\mathrm{delay}} \sim \mathcal{U}(0,0.5)\,\mathrm{s}$, and a random speed perturbation of $\pm 0.1875\%$ of the total tendon length per second. Each simulation trial is $10$ s long, during which tendons are shortened a total of $30\%$ of their original length ($\gamma=0.3$). All simulations used a time step of $20$ $\mu$s. Simulations were performed on the Virginia Tech Advanced Research Computing Owl cluster consisting of water-cooled 96-core AMD EPYC 9454 (Genoa) processors. The average single-core runtime per trial is approximately $1.2$ hours.

\textbf{Computation of link-based quantities.}
To compute link-based quantities, we consider each tentacle as a discretized, directed curve defined by the open axial curve ${\mathbf{x}_i}(s,t)$ and an associated normal vector ${\mathbf{n}}_{i}(s,t)$.  We extend this curve by appending arbitrarily long, straight, untwisted segments to the base and tip of the arm.  From this point, ${\mathbf{x}}_i(s,t)$ will refer to this extended curve.

The linking number is then numerically evaluated using Gauss integrals \cite{klenin2000computation}, which represent the average number of crossings from spherical projectile views.  We compute link $\link$ according to the methodology developed by Charles et. al.  \cite{Charles:2019} for knot theory calculations of open curves. For two arbitrary curves $\mathbf{u}_1$, and $\mathbf{u}_2$, we compute link as
\begin{equation}
    \link(\mathbf{u}_1,\mathbf{u}_2) = \sum_{k=0}^{n+1} \sum_{l=0}^{n+1} \frac{1}{4\pi} \Omega_{k,l}
\end{equation}
where $\Omega_{k,l}$ is the solid angle determined by curve segments ${\mathbf{x}}_i^k$ and ${\mathbf{x}}_j^l$ (e.g. the $k$-th element of curve $\mathbf{x}_i(s,t)$ and the $l$-th element of curve $\mathbf{x}_j(s,t)$). 
The sum corresponds to $n$ physical segments and two additional segments. 

We define three link-based quantities. For a single tentacle, self-link $\link_s^{i}=\link(\mathbf{x}_i,\mathbf{a}_i)$, is computed using the centerline curve of a tentacle $\mathbf{x}_i$ and an auxiliary curve slightly offset by a parameter $\alpha$ from the central curve $a_i = \mathbf{x}_i+\alpha \mathbf{n}_i$ in the local normal direction $\mathbf{n}_i$.Mutual-link $\link_m^{i,j}=\link(\mathbf{x}_i,\mathbf{x}_j)$ of two tentacles is computed based on the two tentacle's centerline curves. From here, the total mutual link of the system is defined as the averaged absolute pairwise linking number
\begin{equation}
    \link_{mutal}=\frac{2}{N(N-1)}\sum^N_{i=1}\sum^N_{j=i+1} |\link_m^{i,j}|
\end{equation}
where $N$ is the number of tentacles. The absolute value is used because we are concerned about the net number of crossings, rather than their sign, which encodes layering order. Finally, object link is defined as the linking number of a tentacle and a straight line aligned with the axial direction vector $\mathbf{c}$ of the target cylinder object
\begin{equation}
    \link_{obj}=\frac{1}{N}\sum^N_{i=1} |\link_o^i (\mathbf{x}_i,\mathbf{c})|.
\end{equation}

\subsection*{Experimental setup}

\textbf{In Air.} 
The in-air experimental setup consists of a rigid metal frame, tendon-routing hardware, an actuation module, and the origami tentacle gripper. The gripper carrier is fixed to the frame using PLA mounting brackets that are bolted in place. Two continuous-rotation servos (peak stall torque 35 kg $\cdot$ cm, operating voltage 8.4~V) drive PLA spools that wind/unwind 40lb fish lines at 60 \%\ of their maximum speed, with a push-button interface used to toggle pull and release directions.

For each trial, the target object is suspended by a fishing line of prescribed length, tied between the object and the carrier center to set the initial distance. The tentacles are initialized in the relaxed configuration, and tendon retraction is applied until the grasp outcome (success or failure) is determined. After each trial, release was triggered by reversing the servo rotation to unwind the tendon from the spool. The tendon was not actively pushed in compression, and no spring-loaded spool or separate slack-management mechanism was used in the current prototype. As the tendon tension decreased, the elastic recovery of the folded Mylar tentacles drove the system back toward their relaxed configurations, allowing the coiling- and tangling-induced object links to disengage.

\textbf{Underwater.} 
The underwater experimental setup consists of a rigid metal frame, 3D-printed mounting hardware, and the tentacle gripper system. The gripper carrier is fixed to the frame using a PLA-CF mount, and the tentacles are manually actuated by pulling the bundled tendons.

For grasping trials, the target object is suspended by a fishing line at a prescribed height relative to the carrier. Each trial begins with the tentacles settled in the relaxed configuration, followed by tendon retraction to execute the grasp.

\medskip
\textbf{Supporting Information} \par
Supporting Information is available from the Wiley Online Library or from the author.

\medskip
\textbf{Acknowledgements} \par 
N.N. and B.Z. acknowledge the support from the National Science Foundation (DCSD-\#2528850). A.B., Z.Z., and S.L. acknowledge the support from the National Science Foundation (FRR-\#2312422). The authors also thank Dr. Austin Phoenix at Virginia Tech NSI for fruitful discussions. 

\medskip

\bibliographystyle{ieeetr}
\bibliography{reference}



\end{document}